\documentclass[11pt,a4paper]{article}
\usepackage[hyperref]{acl2019}
\usepackage{times}
\usepackage{latexsym}

\usepackage{amssymb}% http://ctan.org/pkg/amssymb
\usepackage{amsmath}
\usepackage{pifont}% http://ctan.org/pkg/pifont
\usepackage{booktabs}

\newcommand{\w}[1]{\emph{#1}}%
\newcommand{\ds}[1]{\textsc{#1}}%

\usepackage{url}
\usepackage{bm}

\usepackage{multirow}
\usepackage{CJKutf8}
\usepackage{subcaption}

\usepackage{graphicx}
\usepackage{tikz}
\usetikzlibrary{matrix,arrows,decorations.pathmorphing}

\usepackage{array,booktabs,makecell,tablefootnote}
\usepackage[para]{footmisc}
\usepackage{threeparttable}

\aclfinalcopy % Uncomment this line for the final submission
\def\aclpaperid{1964} %  Enter the acl Paper ID here
\newcommand*\samethanks[1][\value{footnote}]{\footnotemark[#1]}

\title{DisSent: Learning Sentence Representations from Explicit Discourse Relations}

\author{Allen Nie\thanks{~equal contribution}~$^1$ ~~ Erin D. Bennett\samethanks~$^{2}$ ~~ Noah D. Goodman$^{1,2}$ \\
  $^1$Department of Computer Science ~~ $^2$Department of Psychology \\
  Stanford University \\
  {\small \tt anie@cs.stanford.edu ~~ \{erindb,ngoodman\}@stanford.edu} \\
}
\date{}

\begin{document}
\maketitle

\begin{abstract}
Learning effective representations of sentences is one of the core missions of natural language understanding.
Existing models either train on a vast amount of text, or require costly, manually curated sentence relation datasets.
We show that with dependency parsing and rule-based rubrics, we can curate a high quality sentence relation task by leveraging explicit discourse relations.
We show that our curated dataset provides an excellent signal for learning vector representations of sentence meaning, representing relations that can only be determined when the meanings of two sentences are combined.
We demonstrate that the automatically curated corpus allows a bidirectional LSTM sentence encoder to yield high quality sentence embeddings and can serve as a supervised fine-tuning dataset for larger models such as BERT.
Our fixed sentence embeddings achieve high performance
on a variety of transfer tasks, including SentEval, and
we achieve state-of-the-art results on Penn Discourse Treebank's implicit relation prediction task.
\end{abstract}

\section{Introduction}

Developing general models to represent the meaning of a sentence is a key task in natural language understanding.
The applications of such general-purpose representations of sentence meaning are many --- paraphrase detection, summarization, knowledge-base population, question-answering, automatic message forwarding, and metaphoric language, to name a few.

We propose to leverage a high-level relationship between sentences that is both frequently and systematically marked in natural language: the \emph{discourse relations} between sentences.
Human writers naturally use a small set of very common transition words between sentences (or sentence-like phrases) to identify the relations between adjacent ideas. 
These words, such as \w{because}, \w{but}, \w{and}, which mark the conceptual relationship between two sentences,
have been widely studied in linguistics, both formally and computationally, and have many different names. We use the name ``discourse markers".

Learning flexible meaning representations requires a sufficiently demanding, yet tractable, training task. Discourse markers annotate deep conceptual relations between sentences. Learning to predict them may thus represent a strong training task for sentence meanings. This task is an interesting intermediary between two recent trends. On the one hand, models like InferSent \cite{conneau2017supervised} are trained to predict entailment---a strong conceptual relation that must be hand annotated. On the other hand, models like BERT \cite{devlin2018bert} are trained to predict random missing words in very large corpora (see Table~\ref{table:data-size} for the data requirements of the models we compare). Discourse marker prediction may permit learning from relatively little data, like entailment, but can rely on naturally occurring data rather than hand annotation, like word-prediction.

We thus propose the DisSent task, which uses the Discourse Prediction Task to train sentence embeddings.
Using a data preprocessing procedure based on dependency parsing, we are able to
automatically curate a sizable training set of sentence pairs.
We then train a sentence encoding model to learn embeddings for each sentence in a pair such that a classifier can identify, based on the embeddings, which discourse marker was used to link the sentences.
We also use the DisSent task to fine-tune larger pre-trained models such as BERT.

We evaluate our sentence embedding model's performance on the standard fixed embedding evaluation framework developed by~\citet{conneau2017supervised}, where during evaluation, the sentence embedding model's weights are not updated. We further evaluate both the DisSent model and a BERT model fine-tuned on DisSent on two classification tasks from the Penn Discourse Treebank (PDTB)~\cite{rashmi2008penn}.

We demonstrate 
that the resulting DisSent embeddings achieve comparable results to InferSent on some evaluation tasks, and superior on others.
The BERT model fine-tuned on the DisSent tasks achieved state-of-the-art on PDTB classification tasks compared to other fine-tuning strategies.

\section{Discourse Prediction Task}

\citet{hobbs1985on} argues that discourse relations are always present, that they fall under a small set of categories, and that they compose into parsable structures.
We draw inspiration from Rhetorical Structure Theory (RST) \cite{mann1988rhetorical}, which deals with the general task of segmenting natural text into elementary discourse units (EDUs) \cite{carlson2001discourse} and parsing into complex discourse structures (e.g. \citealt{lin2019unified}).
However, for our task, we narrow our scope to a small subset of especially straightforward discourse relations.
First, we restrict our interest to only a subset of EDUs (sentence-like text fragments) that can be interpreted as grammatically complete sentences in isolation.
This includes EDUs that appear as full sentences in the original text, as well as subordinate clauses with overt subjects and finite verb phrases.
Second, we focus here on explicit discourse markers between adjacent sentences (or EDUs), rather than implicit relations between a sentence (or EDU) and the related discourse.
This is a significant simplification from related work in discourse theory, e.g.
describing the wealth of complex structures a discourse can take \cite{webber2003anaphora}
or compiling a comprehensive set of discourse relations \cite{rashmi2008penn,hobbs1990literature,hobbs1985on,jasinskaja2015rhetorical,knott1996data}.
We are able to make this simplification because our goal is not to annotate natural text, but to curate a set of sentence pairs for a particular set of discourse relations.

\begin{table}[t]
\centering
\footnotesize
\begin{tabular}{ c c c }
\toprule
Task & \# of examples & \# of tokens \\
\midrule
SNLI + MNLI & 0.9M & 16.3M  \\
DisSent Books 5 & 3.2M & 63.5M \\
SkipThought & --- & 800M \\
BERT MLM/NSP & --- & 3300M \\
\bottomrule
\end{tabular}
\caption{Training data size (in millions) in each pre-training task. DisSent Books 5 only uses 5 discourse markers instead of all.}
 \label{table:data-size}
\end{table}

\begin{table}[htb]
\centering
\footnotesize
\begin{tabular}{ c c r c c r c }
 \toprule
 Marker & %Frequency (in 100K) &
 \multicolumn{3}{c}{Extracted Pairs} & \multicolumn{3}{c}{Percent (\%)} \\
 \midrule
but  & & 1,028,995  & &  &  21.86 \\
and & & 1,020,316  & & & 21.68 \\
as  & &  748,886   & &  & 15.91 \\
when  & &  527,031  & &  &  11.20 \\
if  & &  472,852  & &  &  10.05  \\
before  & &  218,305   & &  &  4.64 \\
because  & &  167,358   & &  &  3.56 \\
while  & &  161,818   & &  &  3.44 \\
though  & &  104,218   & &  &  2.21 \\
after  & &   95,847   & &  &  2.04 \\
so   & &  76,940   & &  &  1.63 \\
although  & &   37,511  & &  &   0.80 \\
then   & &  16,429  & &  &   0.35 \\
also   & &  16,365   & &  &  0.35 \\
still  & &   13,421   & &  &  0.29 \\
 \midrule
 Total & & 4,706,292 & & & 100.0 \\
\bottomrule
\end{tabular}
\caption{Number of pairs of sentences extracted from BookCorpus for each discourse marker and percent of each marker in the resulting dataset.}
 \label{table:frequencies}
\end{table}

With this focus in mind, we propose a new task for natural language understanding: discourse marker prediction.
Given two sentences in our curated corpus (which may have been full sentences in the original text or may have been subclauses), the model must predict which discourse marker was used by the author to link the two ideas.
For example, ``She's late to class $\rule{1cm}{0.15mm}$ she missed the bus'' would likely be completed with \w{because}, but ``She's sick at home $\rule{1cm}{0.15mm}$ she missed the class'' would likely be completed with \w{so}, and ``She's good at soccer $\rule{1cm}{0.15mm}$ she missed the goal'' would likely be completed with \w{but}.
%All of t
These pairs have similar syntactic structures and many words in common, but the meanings of the component sentences lead to strong intuitions about which discourse marker makes the most sense.
Without a semantic understanding of the sentences, we would not be able to guess the correct relation.
We argue that success at choosing the correct discourse marker requires a representation that reflects the full meaning of a sentence.

We note that perfect performance at this task is impossible for humans \cite{malmi-etal-2018-automatic}, because different discourse markers can easily appear in the same context.
For example, in some cases, markers are (at least close to) synonymous with one another \cite{knott1996data}.
Other times, it is possible for multiple discourse markers to link the same pair of sentences and change the interpretation. (In the sentence ``Bob saw Alice was at the party, (then$|$so$|$but) he went home,'' changing the discourse marker drastically changes our interpretation of Bob's goals and feeling towards Alice.)
Despite this ceiling on absolute performance, a discourse marker can frequently be inferred from the meanings of the sentences it connects, making this a useful training task.

\section{Model}
\label{model}

\subsection{Sentence Encoder Model}

%\paragraph{DisSent Model} 
We adapt the best architecture from Conneau et al. \shortcite{conneau2017supervised} as our sentence encoder.
This architecture uses a standard bidirectional LSTM \cite{graves2013hybrid}, followed by temporal max-pooling to create sentence vectors.
We parameterize the BiLSTM with the different weights $\theta_1$ and $\theta_2$ to reflect the asymmetry of sentence processing.
% Allen: changed last sentence.
We then concatenate the forward and backward encodings.

We apply global max pooling %on the resulting vectors
to construct the encoding for each sentence.
That is, we apply an element-wise max operation over the temporal dimension of the hidden states. %, resulting in a single vector with dimensionality $2d$.
Global max pooling builds a sentence representation from all time steps in the processing of a sentence \cite{collobert2008unified,conneau2017supervised}, providing regularization and shorter back-propagation paths.

\begin{equation}
\begin{split}
    \overrightarrow{h_t} &= \text{LSTM}_t(w_1, ..., w_t | \theta_1) \\
    \overleftarrow{h_t} &= \text{LSTM}_t(w_T, ..., w_t | \theta_2) \\
    h_t &= [\overrightarrow{h_t};\overleftarrow{h_t}] \\
    % \mathbf{H} &= [h_0, ..., h_T] \\
    % s^i &= \mbox{max}(h^i_0, ..., h^i_T) \mbox{ for $i=1, ..., d$} \\
% \end{split}
% \label{eq:lstm}
% \end{equation}
% \begin{equation}
% \begin{split}
    s_i &= \text{MaxPool} (h_1, ..., h_T) \\
\end{split}
\label{eq:temp_max}
\end{equation}

% % After pooling, we concatenate the final forward and backward encodings into a single encoding for each sentence via addition, $s_i{=} [\overrightarrow{s_i}; \overleftarrow{s_i}]$. 
% \ndg{addition of concatenation?}
% % Allen: edited last sentence

% \erindb{reviewer said this paragraph was unclear}
Our objective is to predict the discourse relations between two sentences from their vectors, $s_i$ where $i \in \{1, 2\}$.
Because we want generally useful sentence vectors after training, the learned computation should happen before the sentences are combined to make a prediction.
However, some non-linear 
interactions between the sentence vectors are likely to be needed.
To achieve this,
% without taking on too much of the work from the embeddings,
we include a fixed set of common pair-wise vector operations: subtraction, multiplication, and average.
\begin{equation}
\begin{split}
    s_{\text{avg}} &= \frac{1}{2} (s_1 + s_2) \\
    s_{\text{sub}} &= s_1 - s_2 \\
    s_{\text{mul}} &= s_1 * s_2 \\
    S = [s_1, &s_2, s_{\text{avg}}, s_{\text{sub}}, s_{\text{mul}}]
\end{split}
\label{eq:vec_op}
\end{equation}

Finally we use an affine fully-connected layer to project the concatenated vector $S$ down to a lower dimensional representation, and then project it down to a vector of label size (the number of discourse markers). We use softmax to compute the probability distribution over discourse relations.

% TODO Allen: maybe comment on why we didn't use other models

\subsection{Fine-tuning Model}

Sentence relations datasets can be used to provide high-level training signals to fine-tune other sentence embedding models. 
In this work, we fine-tune BERT~\cite{devlin2018bert} on the DisSent task and evaluate its performance on the PDTB %Penn Discourse Treebank
implicit relation prediction task. We use the BERT-base model which has a 12-layer Transformer encoder. We directly use the $\texttt{[CLS]}$ token's position as the embedding for the entire sentence pair.

After training BERT-base model on the DisSent task, we continue to fine-tune BERT-base model on other evaluation tasks to see if training on DisSent tasks provides additional performance improvement and learning signal for the BERT-base model.

\section{Data Collection}

\begin{table*}
    \centering
    \footnotesize
    \begin{tabular}{l|c|l}
        \toprule
        \multicolumn{1}{c|}{S1} & marker & \multicolumn{1}{c}{S2}  \\
        \midrule

Her eyes flew up to his face.
&and&
Suddenly she realized why he looked so different. \\

The concept is simple.
&but&
The execution will be incredibly dangerous. \\

You used to feel pride.
&because&
You defended innocent people. \\

Ill tell you about it.
&if&
You give me your number. \\

Belter was still hard at work.
&when&
Drade and barney strolled in. \\

We plugged bulky headsets into the dashboard.
&so&
We could hear each other when we spoke into the microphones. \\

It was mere minutes or hours.
&before&
He finally fell into unconsciousness. \\

And then the cloudy darkness lifted.
&though&
The lifeboat did not slow down. \\
    \bottomrule
    \end{tabular}
    \caption{\textbf{Example pairs} from our Books 8 dataset.}
    \label{tab:task_examples}
\end{table*}

We present an automatic way to collect a large dataset of sentence pairs and the relations between them from natural text corpora using a set of explicit discourse markers and universal dependency parsing \cite{schuster2016enhanced}.

\subsection{Corpus and Discourse Marker Set} 

For training and evaluation datasets, we collect sentence pairs from
BookCorpus \cite{zhu2015aligning}, text from unpublished novels (\emph{Romance}, \emph{Fantasy}, \emph{Science fiction}, and \emph{Teen} genres), which was used by \citet{kiros2015skip} to train their SkipThought model.
We identified common discourse markers, choosing those with a frequency greater than 1\% in PDTB. Our final set of discourse markers is shown in Table~\ref{table:frequencies} and we experiment with three subsets of discourse markers (\ds{ALL}, 5, and 8), shown in Table~\ref{table:marker_sets}.

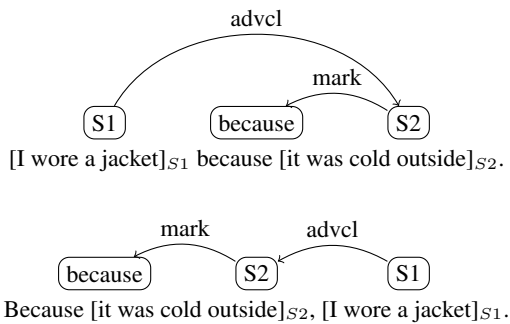
\begin{figure}[t]
    \centering
    \footnotesize
    % \begin{subfigure}{0.5\textwidth}
        \begin{tikzpicture}
            \node (S1)[shape=rectangle,draw=black,rounded corners]
                at (0, 2.5) {S1};
            \node (because)[shape=rectangle,draw=black,rounded corners]
                at (2, 2.5) {because};
            \node (S2)[shape=rectangle,draw=black,rounded corners]
                at (4, 2.5) {S2};
            \node (sentence)
                at (2, 2) {[I wore a jacket]$_{S1}$ because [it was cold outside]$_{S2}$.};

            \path [->] (S1) edge[bend left=60]
                node[above] {advcl} (S2);
            \path [->] (S2) edge[bend right=30]
                node[above] {mark} (because); 
    %     \end{tikzpicture}
    %     \caption{one version}
    %     \label{S1becauseS2}
    % \end{subfigure}
    % \begin{subfigure}{0.5\textwidth}
    %     \begin{tikzpicture}
            \node (S1)[shape=rectangle,draw=black,rounded corners]
                at (4, 0.5) {S1};
            \node (because)[shape=rectangle,draw=black,rounded corners]
                at (0, 0.5) {because};
            \node (S2)[shape=rectangle,draw=black,rounded corners]
                at (2, 0.5) {S2};
            \node (sentence)
                at (2, 0) {Because [it was cold outside]$_{S2}$, [I wore a jacket]$_{S1}$.};

            \path [->] (S1) edge[bend right=30]
                node[above] {advcl} (S2);
            \path [->] (S2) edge[bend right=30]
                node[above] {mark} (because); 
        \end{tikzpicture}
    %     \caption{another version}
    %     \label{becauseS2S1}
    % \end{subfigure}
    \caption{\textbf{Dependency patters for extraction:} While the relative order of a discourse marker (e.g. \w{because}) and its connected sentences is flexible, the dependency relations between these components within the overall sentence remains constant. See Appendix~\ref{sec:dependency_details} for dependency patterns for other discourse markers.
    }
    \label{fig:becausedependencies}
\end{figure}

\subsection{Dependency Parsing} 

Many discourse markers in English occur almost exclusively between the two statements they connect, 
but for other discourse markers, their position relative to their connected statements can vary
(e.g. Figure~\ref{fig:becausedependencies}).
For this reason, we use the Stanford CoreNLP dependency parser \cite{schuster2016enhanced} to 
extract the appropriate pairs of sentences (or sentence-like EDUs) for a discourse marker, in the appropriate conceptual order.
Each discourse marker, when it is used to link two statements, is parsed by the dependency parser in a systematic way, though  different discourse markers may have different corresponding dependency patterns linking them to their statement pairs.\footnote{See Appendix~\ref{sec:dependency_details} for more details on dependency-based extraction.}

Within the dependency parse, we search for the governor phrase (which we call ``S2'') of the discourse marker and check for the appropriate dependency relation.
If we find no such phrase, we reject the example entirely (thus filtering out polysemous usages, like ``that's \emph{so} cool!'' for the discourse marker \w{so}).
If we find such an S2, we search for ``S1'' within the same sentence (SS).
Searching for this relation allows us to capture pairs where the discourse marker starts the sentence and connects the following two clauses (e.g. ``\w{Because} [it was cold outside]$_{S2}$, [I wore a jacket]$_{S1}$.'').
If a sentence in the corpus contains only a discourse marker and S2,
we assume the discourse marker links to the immediately previous sentence (IPS), which we label S1.

For some markers, we further filter based on the order of the sentences in the original text.
For example, the discourse marker \w{then} always appears in the order "S1, then S2", unlike \w{because}, which can also appear in the order "Because S2, S1".
Excluding proposed extractions in an incorrect order makes our method more robust to incorrect dependency parses.

\subsection{Training Dataset}

Using these methods, we curated a dataset of 4,706,292 pairs of sentences for 15 discourse markers.
Examples are shown in Table~\ref{tab:task_examples}.
We randomly divide the dataset into train/validation/test set with 0.9, 0.05, 0.05 split. The dataset is inherently unbalanced, but 
the model is still able to learn rarer classes quite well (see Appendix~\ref{sec:balanced_confusion} for more details on the effects of class frequencies).
Our data are publicly available\footnote{\url{https://github.com/windweller/DisExtract}}.

\section{Related Work}

Current state of the art models 
either rely on completely supervised learning through high-level classification tasks or unsupervised learning.

% unsupervised 
% supervised

Supervised learning has been shown to yield general-purpose representations of meaning, training on semantic relation tasks like Stanford Natural Language Inference (SNLI) and MultiNLI \cite{bowman2015large,williams2017broad,conneau2017supervised}.
Large scale joint supervised training has also been explored by \citet{subramanian2018learning}, who trained a sentence encoding model on five language-related tasks.
These supervised learning tasks often require human annotations on a large amount of data which are costly to obtain.
Our discourse prediction approach extends these results in that we train on semantic relations, but we use dependency patterns to automatically curate a sizable dataset.

In an unsupervised learning setting, SkipThought \cite{kiros2015skip} learns a conditional joint probability distribution for the next sentence. ELMo~\cite{peters2018deep} uses a BiLSTM to predict the missing word using the masked language modeling (MLM) objective. OpenAI-GPT2~\cite{radford2019language} directly predicts the next word. BERT~\cite{devlin2018bert} uses MLM as well as predicting whether the next sequence comes from the same document or not. Despite the overwhelming success of these models, \citet{phang2018sentence} shows that fine-tuning these models on supervised learning datasets can yield improved performance over difficult natural language understanding tasks.

\citet{jernite_discourse-based_2017} have proposed a model that also leverages discourse relations.
They manually categorize discourse markers based on human interpretations of discourse marker similarity, and the model predicts the category instead of the individual discourse marker.
Their model also trains on auxiliary tasks, such as sentence ordering and ranking of the following sentence and must compensate for data imbalance across tasks.
Their data collection methods only allow them to look at paragraphs longer than 8 sentences, and sentence pairs with sentence-initial discourse markers, resulting in only 1.4M sentence pairs from a much larger corpus.
Our proposed model extracts a wider variety of sentence pairs, can be applied to corpora with shorter paragraphs, and includes no auxiliary tasks.

\section{Experiments}

For all our models, we tuned hyperparameters on the validation set, and report results from the test set. We use stochastic gradient descent with initial learning rate 0.1, and anneal by the factor of 5 each time validation accuracy is lower than in the previous epoch. We train our fixed sentence encoder model for 20 epochs, and use early stopping to prevent overfitting. We also clip the gradient norm to 5.0. 
We did not use dropout in the fully connected layer in the final results because our initial experiments with dropout showed lower performance when generalizing to SentEval. We experimented with both global mean pooling and global max pooling and found the later to perform much better at generalization tasks.
All models we report used a 4096 hidden state size. 
We are able to fit our model on a single Nvidia Titan X GPU.

\paragraph{Fine-tuning} We fine-tune the BERT-base model on the DisSent tasks with 2e-5 learning rate for 1 epoch because all DisSent tasks corpora are quite large and fine-tuning for longer epochs did not yield improvement. We fine-tune BERT on other supervised learning datasets for multiple epochs and select the epoch that provides the best performance on the evaluation task. We find that fine-tuning on MNLI is better than on SNLI or both combined. This phenomenon is also discussed in \citet{phang2018sentence}.

% 1. Add Fine-tuning details!!!

\paragraph{Discourse Marker Set}

We experimented with three subsets of discourse markers, shown in Table~\ref{table:marker_sets}.
We first trained over all of the discourse markers in our \ds{ALL} marker set. The model achieved 67.5\% test accuracy on this classification task. Overall we found that markers with similar meanings tended to be confusable with one another. A more detailed analysis of the model's performance on this classification task is presented in Appendix \ref{sec:balanced_confusion}.

Because there appears to be intrinsic conceptual overlap in the set of \ds{ALL} markers, we experimented on different subsets of discourse markers. 
We choose sets of 5 and 8 discourse markers that were both non-overlapping and frequent.
The set of sentence pairs for each smaller dataset is a strict subset of those in any larger dataset.
Our chosen sets are shown in Table~\ref{table:marker_sets}.

\begin{table}[t]
    \centering
    \footnotesize
    \begin{tabular}{l|l}
        \toprule
        Label & Discourse Markers \\
        \midrule
        Books 5  & \w{and}, \w{but}, \w{because}, \w{if}, \w{when} \\
        Books 8 & \w{and}, \w{but}, \w{because}, \w{if}, \w{when}, \w{before}, \\ & \w{though}, \w{so} \\
        Books ALL & \w{and}, \w{but}, \w{because}, \w{if}, \w{when}, \w{before}, \\ & \w{though}, \w{so}, \w{as}, \w{while}, \w{after}, \w{still}, \w{also}, \\ & \w{then}, \w{although} \\
        \bottomrule  % \ds{ALL}
    \end{tabular}
    \caption{\textbf{Discourse marker sets} used in our experiments. Books ALL contains 4.7M sentence pairs, Books 8 contains 3.6M, and Books 5 contains 3.2M.}
    \label{table:marker_sets}
\end{table}

\paragraph{Marked vs Unmarked Prediction Task}

Adjacent sentences will always have a relationship,
but some are marked with discourse markers while others are not.
Humans have been shown to perform well above chance at guessing whether a discourse marker is marked vs. unmarked \cite{patterson2013predicting,yung2017psycholinguistic}, indicating a systematicity to this decision.

We predict that high quality sentence embeddings will contain useful information to determine whether a discourse relation is explicitly marked.
Furthermore, success at this task could help natural language generation models to generate more human-like long sequences.

To test this prediction, we create an additional set of tasks based on
Penn Discourse Treebank \cite{rashmi2008penn}.
This hand-annotated dataset contains expert discourse relation annotations between sentences.
We collected 34,512 sentences from PDTB\footnote{https://github.com/cgpotts/pdtb2} (see Appendix), where 16,224 sentences are marked with implicit relation type, and 18,459 are marked with explicit relation type.

\paragraph{Implicit Relation Prediction Task}

\citet{sporleder2008using} have argued that sentence pairs with explicitly marked relations are qualitatively different from those where the relation is left implicit.
However, despite such differences,
\citet{qin2017adversarial} were able to use an adversarial network to leverage
explicit discourse data as additional training to increase the performance on the implicit discourse relation prediction task. 
We use the same dataset split scheme for this task as for the implicit vs explicit task discussed above. Following \citet{ji2014one} and \citet{qin2017adversarial}, we predict the 11 most frequent relations. 
There are 13,445 pairs for training, and 1,188 pairs for evaluation.

\paragraph{SentEval Tasks} \label{sec:generalization}
We evaluate the performance of generated sentence embeddings from our fixed sentence encoder model on a series of natural language understanding benchmark tests provided by \citet{conneau2017supervised}.
The tasks we chose include sentiment analysis (MR, SST), question-type (TREC), product reviews (CR), subjectivity-objectivity (SUBJ), opinion polarity (MPQA), entailment (SICK-E), relatedness (SICK-R), and paraphrase detection (MRPC).
These are all classification tasks with 2-6 classes, except for relatedness, for which the model predicts human similarity judgments.

\subsection{Results}

\begin{table}[htb]
\centering
\footnotesize
% \begin{tabular}{ c | @{\hskip 0.2in} c @{\hskip 0.3in} c @{\hskip 0.3in} c}
\begin{tabular}{ c | c c c c c c}
\toprule
 & \multicolumn{2}{c}{All} & \multicolumn{2}{c}{Books 8} & \multicolumn{2}{c}{Books 5} \\
Model & F1 & Acc & F1 & Acc & F1 & Acc \\
\midrule 
% DisSent & 66.7 & 68.0 & 73.6 & 73.3 & 77.5 & 77.4 \\ 
GloVe-bow & 17.1 & 41.8 & 27.6 & 47.3 & 41.7 & 52.5 \\
Ngram-bow & 28.1 & 51.8 & 44.0 & 58.1 & 54.1 & 63.3 \\
BiLSTM & 47.2 & 67.5 & 64.4 & 73.5 & 72.1 & 77.3 \\ 
BERT & 60.1 & 77.5 & 76.2 & 82.9 & 82.6 & 86.1 \\
% \midrule
% \midrule
\bottomrule
\end{tabular}
\caption{\textbf{Discourse classification task performance:}  Unweighted average F1 across discourse markers on the test set, and overall accuracy. Ngram-bow is a bag-of-words model built on mixture of ngram features. GloVe-bow averages word embedding with correction to frequency~\cite{arora2016simple}. BiLSTM is the DisSent sentence encoder model. BERT is finetuned on all of the DisSent tasks.}
 \label{table:intrinsic_eval}
\end{table}

\paragraph{Training Task}
On the discourse marker prediction task used for training,
we achieve high levels of test performance for all discourse markers.
(Though it is interesting that \emph{because}, perhaps the conceptually deepest relation, is also systematically the hardest for our model.)
The larger the set of discourse markers, the more difficult the task becomes, and we therefore see lower test accuracy despite larger dataset size.
We conjecture that as we increase the number of discourse markers, we also increase the ambiguity between them (semantic overlap in discourse markers' meanings), which may further explain the drop in performance.
The training task performance for each subset is shown in Table \ref{table:intrinsic_eval}. We provide per-discourse-marker performance in the Appendix.

\paragraph{Discourse Marker Set}
Varying the set of discourse markers doesn't seem to help or hinder the model's performance on generalization tasks.
Top generalization performance on the three sets of discourse markers 
is shown in Table \ref{table:extrinsic_eval}.
Similar generalization performance was achieved when training on 5, 8, and all 15 discourse markers.

The similarity in generalization performance across discourse sets shows that the top 
markers capture most relationships in the training data.

\begin{table*}[htb]
\footnotesize
% \begin{minipage}{\textwidth}
\centering {
\begin{tabular}{ c | c c c c c c c c c }
\toprule
Model & MR & CR & SUBJ & MPQA & SST & TREC & SICK-R & SICK-E & MRPC \\ % & DIS
\midrule
\midrule
\multicolumn{10}{c}{Self-supervised training methods} \\
\midrule
DisSent Books 5$^\dagger$ & \underline{80.2} & \underline{85.4} & 93.2 & 90.2 & 82.8 & 91.2 & 0.845 & 83.5 & \underline{76.1} \\ % & 81.5 75.7
DisSent Books 8$^\dagger$ & 79.8 & 85.0 & 93.4 & \underline{90.5} & 83.9 & 93.0 & \underline{0.854} & \underline{83.8} & \underline{76.1} \\ % & 81.8 80.2
DisSent Books \ds{ALL}$^\dagger$ & 80.1 & 84.9 & \underline{93.6} & 90.1 & \underline{84.1} & \textbf{\underline{93.6}} & 0.849 & 83.7 & 75.0 \\ % 79.9 & 85.1 
Disc BiGRU%\footnote{\cite{jernite_discourse-based_2017}}
& {---} & --- &  88.6 & --- & --- & 81.0 & --- & --- & 71.6  \\ % . & ---
\midrule
\multicolumn{10}{c}{Unsupervised training methods} \\
\midrule
FastSent%\footnote{\cite{hill2016learning}}
& 70.8 & 78.4 & 88.7 & 80.6 & --- & 76.8 & --- & --- & 72.2  \\ % & ---
FastSent + AE%\footnote{\cite{hill2016learning}}
& 71.8 & 76.7 & 88.8 & 81.5 & --- & 80.4 & --- & --- & 71.2  \\ % & ---
Skipthought%\footnote{\cite{kiros2015skip}}
& 76.5 & 80.1 & 93.6 & 87.1 & 82.0 & 92.2 & 0.858 & 82.3 & 73.0  \\ % 70.1 & 77.3
Skipthought-LN%\footnote{\cite{conneau2017supervised}} % \footnotemark[4]\footnotemark[5]
& 79.4 & 83.1 & 93.7 & 89.3 & 82.9 & 88.4 & 0.858 & 79.5 & --- \\ % & ---
\midrule
\multicolumn{10}{c}{Supervised training methods} \\
\midrule
DictRep (bow)%\footnote{\cite{conneau2017supervised}}
& 76.7 & 78.7 & 90.7 & 87.2 & --- & 81.0 & --- & --- & ---  \\ % & ---
InferSent%\footnote{\cite{conneau2017supervised}}
& 81.1 & 86.3 & 92.4 & 90.2 & \textbf{84.6} & 88.2 & 0.884 & 86.1 & 76.2 \\ % 65.4 % 71.6  & 72.5
\midrule
\multicolumn{10}{c}{Multi-task training methods} \\
\midrule
LSMTL
% this footnote doesn't show up without the tfn command
& \textbf{82.5} & \textbf{87.7} & \textbf{94.0} & \textbf{90.9} & 83.2 & 93.0 & \textbf{0.888} & \textbf{87.8} & \textbf{78.6} \\ % & ---
\bottomrule
\end{tabular}}
% \caption{\textbf{SentEval Task Results Using Fixed Sentence Encoder.} We report the best results for generalization tasks. $^\dagger$ indicates models that we trained. DisSent uses a BiLSTM encoder with 4096 hidden state dimensions. InferSent \protect\cite{conneau2017supervised} uses 4096 embedding dimensions. Disc BiGRU \protect\cite{jernite_discourse-based_2017} hidden state has 512 dimensions. FastSent and FastSent + AE \protect\cite{hill2016learning} have 500 dimensions. SkipThought \protect\cite{kiros2015skip} and SkipThought-LN \protect\cite{conneau2017supervised} models trained on 600-dimension word embeddings and produced 2400-dimension sentence embeddings. DictRep (bow) is from \protect\citet{conneau2017supervised}. LSMTL \protect\cite{subramanian2018learning} uses 2048-dimension bi-directional GRU as encoder, and trained on 512 dimension word embeddings. Globally best results are shown in \textbf{bold}, best DisSent results are \underline{underlined}.}
\caption{\textbf{SentEval Task Results Using Fixed Sentence Encoder.} We report the best results for generalization tasks. $^\dagger$ indicates models that we trained. FastSent, FastSent + AE~\cite{hill2016learning}, SkipThought~\cite{kiros2015skip}, SkipThought-LN, DictRep (bow), and InferSent are reported from \citet{conneau2017supervised}. LSMTL is reported from \citet{subramanian2018learning}. Globally best results are shown in \textbf{bold}, best DisSent results are \underline{underlined}.}
 \label{table:extrinsic_eval}
 %  
%  \end{minipage}
\end{table*}

\paragraph{Marked vs Unmarked Prediction Task}

In determining whether a discourse relation is marked or unmarked, DisSent models outperform InferSent and SkipThought (as well as previous approaches on this task) by a noticeable margin. Much to our surprise, fine-tuned BERT models are not able to perform better than the BiLSTM sentence encoder model. We leave explorations of this phenomenon to future work.
We report the results in Table~\ref{table:pdtb-implicit} under column MVU.

\paragraph{Implicit Discourse Relation Task} 
Not surprisingly, DisSent task provided the much needed distant supervision to classify the types of implicit discourse relations much better than InferSent and SkipThought. DisSent outperforms word vector models evaluated by \citet{qin2017adversarial}, and is only 3.3\% lower than the complex state of the art model that uses adversarial training designed specifically for this task. 
When we fine-tune BERT models on the DisSent corpora, we are able to outperform all other models and achieve state-of-the-art result on this task.
We report the results in Table~\ref{table:pdtb-implicit} under column IMP.

\begin{table}[htb]
\centering
\footnotesize
\begin{tabular}{c | c c}
\toprule
Model & IMP & MVU \\
\midrule
\multicolumn{3}{c}{Sentence Encoder Models} \\
\midrule
SkipThought \cite{kiros2015skip} & 9.3 & 57.2 \\
InferSent \cite{conneau2017supervised} & 39.3 & 84.5 \\ % 38.4
\citet{patterson2013predicting} & --- & 86.6 \\
\midrule
DisSent Books 5 & 40.7 & 86.5\\
DisSent Books 8 & 41.4 & \textbf{87.9} \\
DisSent Books ALL & \textbf{42.9} & 87.6 \\
\midrule
\multicolumn{3}{c}{Fine-tuned Models} \\
\midrule
BERT & 52.7 & 80.5 \\
BERT + MNLI & 53.7 & 80.7 \\
BERT + SNLI + MNLI & 51.3 & 79.8 \\
% \midrule
BERT + DisSent Books 5 & \textbf{54.7} & 81.6 \\
BERT + DisSent Books 8 & 52.4 & 80.6 \\
BERT + DisSent Books ALL & 53.2 & \textbf{81.8} \\
\midrule 
\multicolumn{3}{c}{Previous Single Task Models} \\
\midrule
Word Vectors \cite{qin2017adversarial} & 36.9 & 74.8 \\
\citet{lin2009recognizing} + Brown Cluster &  40.7 & --- \\
Adversarial Net \citep{qin2017adversarial} & \textbf{46.2} & --- \\
\bottomrule
\end{tabular}
\caption{\textbf{Discourse Generalization Tasks using PDTB:} We report test accuracy for sentence embedding and state-of-the-art models.}
 \label{table:pdtb-implicit}
\end{table}
% Following the metric used in these literature, 
%  for sentence embedding models, as well as baselines and state of the art for these task.

\paragraph{SentEval Tasks}

Results of our models, and comparison to other approaches, are shown in Table \ref{table:extrinsic_eval}.
Despite being a much simpler task than SkipThought and allowing for much more scalable data collection than InferSent, DisSent performs as well or better than these approaches on most generalization tasks.

DisSent and InferSent do well on different sets of tasks.
In particular, DisSent outperforms InferSent on TREC (question-type classification).
InferSent outperforms DisSent on the tasks most similar to its training data, SICK-R and SICK-E.
These tasks, like SNLI, were crowdsourced, and seeded with images from Flickr30k corpus \cite{young2014image}.

Although DisSent is trained on a dataset derived from the same corpus as SkipThought, DisSent almost entirely dominates SkipThought's performance across all tasks.
In particular, on the SICK dataset, DisSent and SkipThought perform similarly on the relatedness task (SICK-R), but DisSent strongly outperforms SkipThought on the entailment task (SICK-E).
This discrepancy highlights an important difference between the two models.
Whereas both models are trained to, given a particular sentence, identify words that appear near that sentence in the corpus,
DisSent focuses on learning specific kinds of relationships between sentences -- ones that humans tend to explicitly mark.
We find that reducing the model's task to only predicting a small set of discourse relations, rather than trying to recover all words in the following sentence, results in better features for identifying entailment and contradiction without losing cues to relatedness.

Overall, on the evaluation tasks we present, DisSent performs on par with previous state-of-the-art models and offers advantages in data collection and training speed.

\section{Extraction Validation}
\label{sec:data-val}

We evaluate our extraction quality by comparing the manually extracted and annotated sentence pairs from Penn Discourse Treebank (PDTB) to our automatic extraction of sentence pairs from the source corpus Penn Treebank (PTB).
On the majority of discourse markers, we can achieve a relatively high extraction precision.

We apply our extraction pipeline on raw PTB dataset because we want to see how well our pipeline converts raw corpus into a dataset. Details of our alignment procedure is described in Appendix~\ref{sec:data_eval}. Overall, even though we cannot construct the explicit discourse prediction section of the PDTB dataset perfectly, training with imprecise extraction has little impact on the sentence encoder model's overall performance.

\begin{figure}[t]
    \centering
    \footnotesize
    \includegraphics[width=0.4\textwidth]{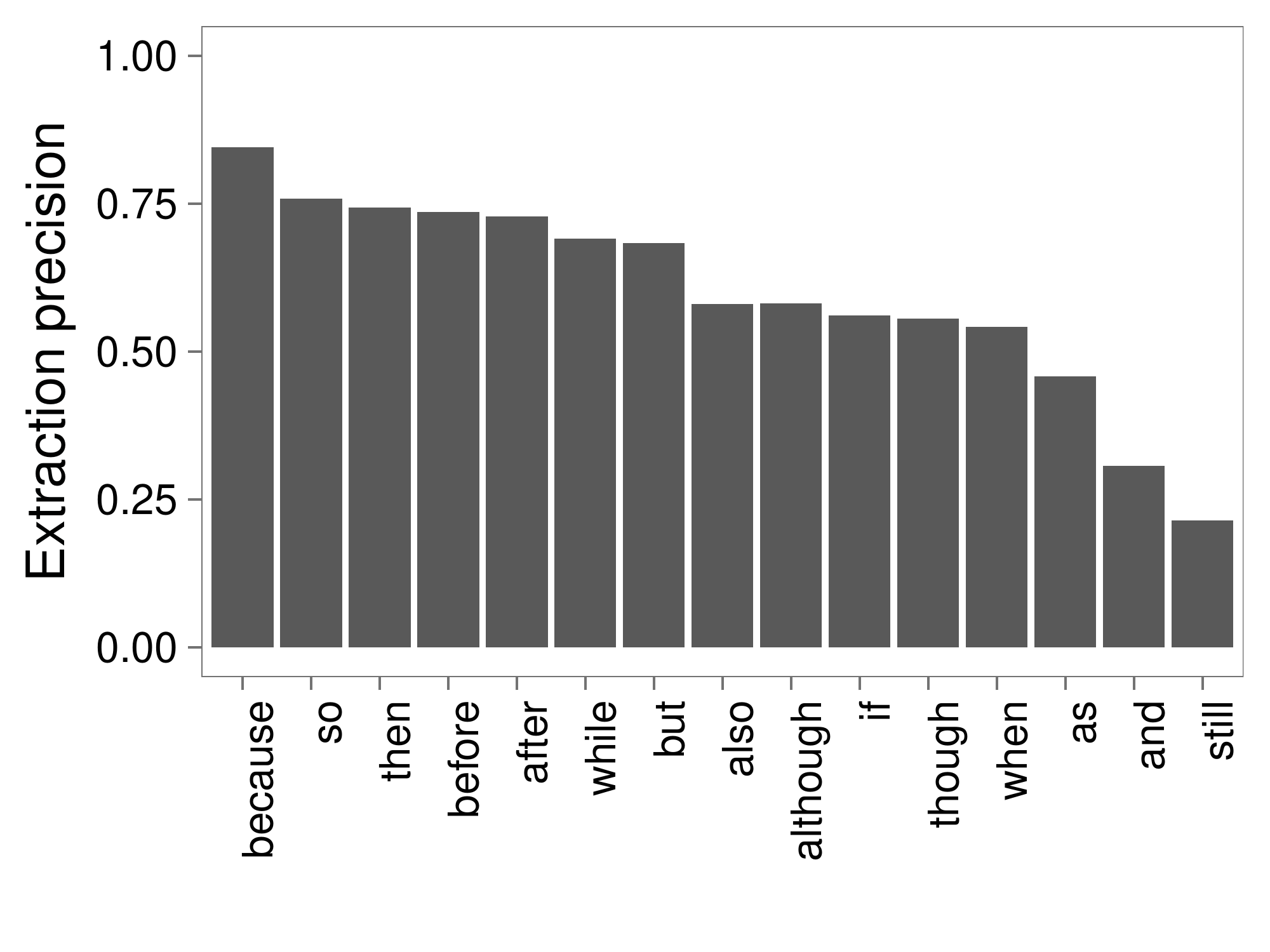}
    \caption{\textbf{Extraction error rates:} proportion of unalignable extracted pairs per discourse marker.}
    \label{fig:extraction_errors}
\end{figure}

We compute the extraction precision as the percentage of PTB extracted pairs that can be successfully aligned to PDTB.
In Figure~\ref{fig:extraction_errors}, we show that extraction precision varies across discourse markers. Some markers have higher quality (e.g. \w{because}, \w{so}) and some lower quality (e.g. \w{and}, \w{still}). 

\begin{figure}[t]
    \centering
    \footnotesize
    \includegraphics[width=0.4\textwidth]{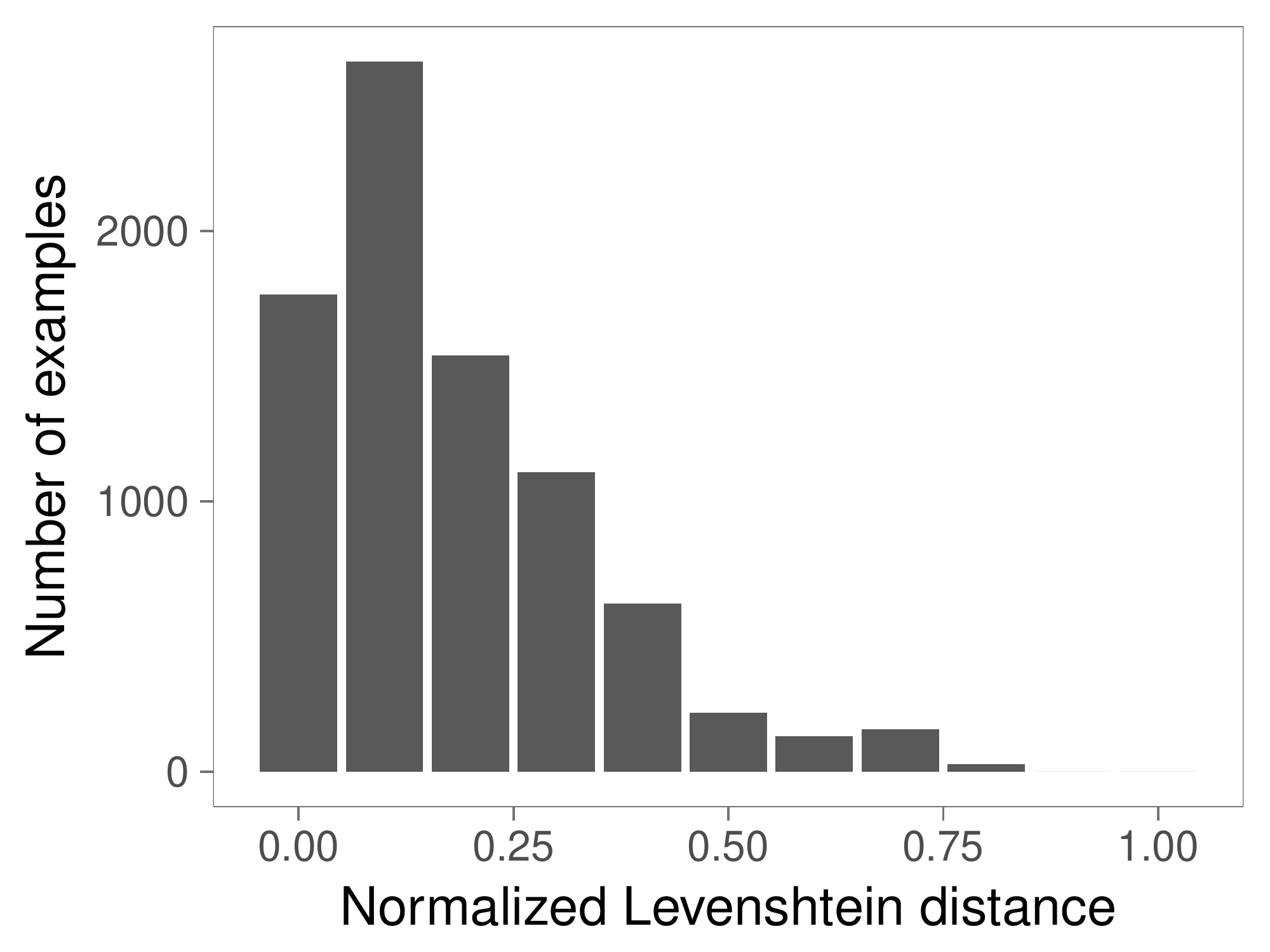}
    \caption{\textbf{Extraction quality for aligned pairs:} Distances from aligned extracted pairs to nearest gold pair.}
    \label{fig:extraction_distances}
\end{figure}

We show in Figure \ref{fig:extraction_distances} that we tend to have low distances overall for the successfully aligned pairs. That is, whenever our extraction pipeline yields a match, the dependency parsing patterns do extract high quality training pairs.

\section{Discussion}

\paragraph{Implicit and explicit discourse relations}

We focus on explicit discourse relations for training our embeddings.
Another meaningful way to exploit discourse relations in training is by leveraging implicit discourse signals.
For instance, \citet{jernite_discourse-based_2017} showed that predicting sentence ordering could help to generate meaningful sentence embeddings.
But adjacent sentences can be related to one another in many different, complicated ways.
For example, sentences linked by contrastive markers, like \w{but} or \w{however} are likely expressing different or opposite ideas.

Identifying other features of natural text that contain informative signals of discourse structure and combining these with explicit discourse markers is an appealing direction for future research. 

\paragraph{Multilingual generalization}

In principle, the DisSent model and extraction methods would apply equally well to multilingual data with minimal language-specific modifications.
Within universal dependency grammar, discourse markers across languages should correspond to structurally similar dependency patterns.
Beyond dependency parsing and minimal marker-specific pattern development (see Appendix~\ref{sec:dependency_details}), our extraction method is automatic, requiring no annotation of the original dataset, and so any large dataset of raw text in a language can be used.

\section{Conclusion}

We present a \emph{discourse marker prediction} task for training and fine-tuning
sentence embedding models.
We train our model on this task and show that the resulting embeddings lead to high 
performance on a number of established tasks for sentence embeddings. We fine-tune larger models on this task and achieve state-of-the-art on the PDTB implicit discourse relation prediction.

A dataset for this task is easy to collect relative to other supervised tasks. It provides cheap and noisy but strong training signals.
Compared to unsupervised methods that train on a full corpus, our method yields more targeted and faster training.
Encouragingly, the model trained on discourse marker prediction achieves comparable generalization performance to other state of the art models.

\section*{Acknowledgement}

We thank Chris Potts for the discussion on Penn Discourse Treebank and the share of preprocessing code. We also thank our anonymous reviewers and the area chair for their thoughtful comments and suggestions.
The research is based upon work supported by the Defense Advanced Research Projects Agency (DARPA), via the Air Force Research Laboratory (AFRL, Grant No. FA8650-18-C-7826). The views and conclusions contained herein are those of the authors and should not be interpreted as necessarily representing the official policies or endorsements, either expressed or implied, of DARPA, the AFRL or the U.S. Government.
The U.S. Government is authorized to reproduce and distribute reprints for Governmental purposes notwithstanding any copyright annotation thereon.   

\bibliography{naaclhlt2019}
\bibliographystyle{acl_natbib}

% \clearpage

\appendix

\section{Appendix}

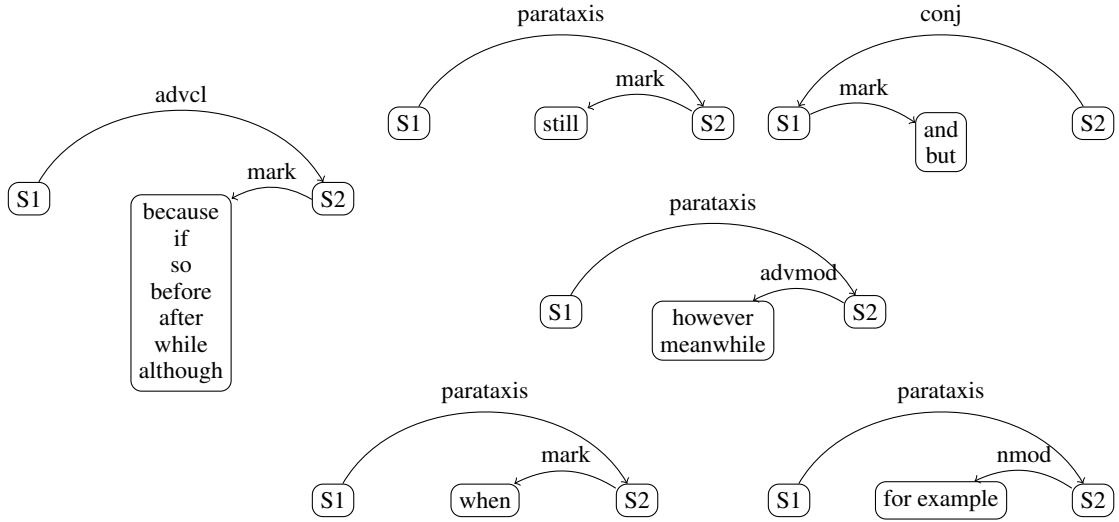
\begin{figure*}[htb]
    \centering
    \footnotesize
        \begin{tikzpicture}
            \node (S1_because)[shape=rectangle,draw=black,rounded corners]
                at (0, 4) {S1};
            \node (because)[shape=rectangle,draw=black,rounded corners,align=center]
                at (2, 2.75) {because \\ if \\ so \\ before \\ after \\ 
                           while \\ although};
            \node (S2_because)[shape=rectangle,draw=black,rounded corners]
                at (4, 4) {S2};

            \path [->] (S1_because) edge[bend left=60]
                node[above] {advcl} (S2_because);
            \path [->] (S2_because) edge[bend right=30]
                node[above] {mark} (because);

            \node (S1_still)[shape=rectangle,draw=black,rounded corners]
                at (5, 5) {S1};
            \node (still)[shape=rectangle,draw=black,rounded corners,align=center]
                at (7, 5) {still};
            \node (S2_still)[shape=rectangle,draw=black,rounded corners]
                at (9, 5) {S2};

            \path [->] (S1_still) edge[bend left=60]
                node[above] {parataxis} (S2_still);
            \path [->] (S2_still) edge[bend right=30]
                node[above] {mark} (still);

            \node (S1_and)[shape=rectangle,draw=black,rounded corners]
                at (10, 5) {S1};
            \node (and)[shape=rectangle,draw=black,rounded corners,align=center]
                at (12, 4.75) {and \\ but};
            \node (S2_and)[shape=rectangle,draw=black,rounded corners]
                at (14, 5) {S2};

            \path [->] (S2_and) edge[bend right=60]
                node[above] {conj} (S1_and);
            \path [->] (S1_and) edge[bend left=30]
                node[above] {mark} (and);

            \node (S1_however)[shape=rectangle,draw=black,rounded corners]
                at (7, 2.5) {S1};
            \node (however)[shape=rectangle,draw=black,rounded corners,align=center]
                at (9, 2.25) {however \\ meanwhile};
            \node (S2_however)[shape=rectangle,draw=black,rounded corners]
                at (11, 2.5) {S2};

            \path [->] (S1_however) edge[bend left=60]
                node[above] {parataxis} (S2_however);
            \path [->] (S2_however) edge[bend right=30]
                node[above] {advmod} (however);

            \node (S1_when)[shape=rectangle,draw=black,rounded corners]
                at (4, 0) {S1};
            \node (when)[shape=rectangle,draw=black,rounded corners,align=center]
                at (6, 0) {when};
            \node (S2_when)[shape=rectangle,draw=black,rounded corners]
                at (8, 0) {S2};

            \path [->] (S1_when) edge[bend left=60]
                node[above] {parataxis} (S2_when);
            \path [->] (S2_when) edge[bend right=30]
                node[above] {mark} (when);

            \node (S1_eg)[shape=rectangle,draw=black,rounded corners]
                at (10, 0) {S1};
            \node (eg)[shape=rectangle,draw=black,rounded corners,align=center]
                at (12, 0) {for example};
            \node (S2_eg)[shape=rectangle,draw=black,rounded corners]
                at (14, 0) {S2};

            \path [->] (S1_eg) edge[bend left=60]
                node[above] {parataxis} (S2_eg);
            \path [->] (S2_eg) edge[bend right=30]
                node[above] {nmod} (eg);

        \end{tikzpicture}
    \caption{Dependency patterns used for extraction for each discourse marker.}
    \label{fig:abstract_dependencies}
\end{figure*}

\subsection{Details on Dependency Based Sentence Extraction\label{sec:dependency_details}}

% We use dependency parsing to identify uses of discourse markers and extract sentence pairs.
While universal dependency grammar provides enough information to identify discourse markers and their connected statements, different discourse markers are parsed with different dependency relations.
For each discourse marker of interest, we identify the appropriate dependency pattern (see Figure~\ref{fig:abstract_dependencies}).
%shows the full set of dependency patterns used by our extraction algorithm for English.
% We further filtered results so that all extracted sentences had at least one verb.

% # if S2 is the whole sentence *and* we're missing S1, let S1 be the previous sentence/

% Although structural similarities exist for different discourse markers in the way universal dependency patterns do not treat every discourse marker the same, 

% Figure~\ref{fig:abstract_dependencies}

We excluded any pair where one of the sentences was less than 5 or more than 50 words long and any pairs where one of the sentences was more than 5 times the length of the other.

Dependency parsing allows us to design our extraction method such that each S1 and S2 is interpretable as a full sentence in isolation, and the appropriate conceptual relation holds between the pair.
However, occasionally we get ungrammatical sentences or the wrong pair of sentences for a relation.
This incorrect extraction can happen in several ways.
% However, despite many advantages, the dependency parser is not able to extract all sentence pairs correctly and can even introduce some problems.%, resulting in ugrammatical sentences or the wrong pair of sentences for a discourse marker. %relations.
% In addition to errors in extraction due to distant relations,
First, we might choose grammatical but incorrect pairs. \citet{rashmi2008penn} found that 61\% of discourse markers appear in the same sentence (SS) with both S1 and S2, and another 30\% link S2 to the immediate predecessor (IPS).
For the remaining examples (non-adjacent previous sentence NAPS - 9\%, or following sentence FS - less than 1\%), our method incorrectly extracts an IPS pair.
Second, not all parses are correct (e.g. ``Himself close his eyes.'' was extracted due to an incorrect parse). 
% (e.g. w non-sentence ``Himself close his eyes.'' was extracted from ``To his shame, he just let himself close his eyes and gave himself over to unconsciousness.'' due to an incorrect parse).
Finally, even with correct parses, some extracted sentences are nonsensical or ungrammatical out of context due to implicit subjects, unresolved pronouns, or marked embedded clauses.
% and other kinds of ellipses.
% Finally, even with correct parses, extraction was imperfect for sentences with implicit, repeated subjects,
%(e.g. ``She was reading her favorite book, [which her sister had given her] \w{when} [she last visited].'').
% (e.g. ``[Wolfe chastised us for not being serious enough]$_{S1}$ \w{and} [gave us high marks for learning the techniques]$_{S2}$'')
% or certain kinds of marked embedded clauses.
% (e.g. ``She was reading her favorite book, [which her sister had given her]$_{S1}$ \w{when} [she last visited]$_{S2}$'').
Fortunately these errors were relatively rare, and many could be avoided simply by enforcing that the extracted sentences each have a main verb and satisfy a minimum length. %In practice, we 
% We therefore excluded any pair where one of the two sentences was less than 5 or more than 50 words long and any pairs where one of the two sentences was more than 5 times the length of the other.
Overall this method extracts high-quality sentence pairs with appropriately labeled relations.

% \subsection{Length-based Filtering}

% As a way to exclude extremely uninformative sentence pairs and standardize lengths of sentences, we filtered pairs based on several criteria on the lengths (in words) of the two sentences.
% We excluded any pair where one of the two sentences was less than 5 or more than 50 words long.
% We additionally excluded any pairs where one of the two sentences was more than 5 times the length of the other.

\subsection{Procedures in Extraction Validation} \label{sec:data_eval}

We preprocess the PTB sentences by limiting the vocabulary size to 10,000 and tokenizing numbers. Then we run our extraction pipeline on the preprocessed PTB. We apply the same preprocessing to the PDTB sentences.

We refer to the gold sentence pair from the PDTB as (G1, G2), and our extracted sentence pair from the PTB as (S1, S2).
We first obtain the minimum of S1-G1 distance and S2-G2 distance over all gold pairs. If this distance is smaller than 0.7, we consider the corresponding gold pair to be an alignment for this extracted pair.

Given an aligned pair ((G1, G2), (S1, S2)), we measure the extraction quality by computing the average of normalized G1-S1 and G2-S2 distance. We compute this distance for all pairs and all discourse markers.

We analyze our extraction quality in two steps: align sentence pairs from the two datasets and then calculate extraction quality on each aligned pair.
In the alignment step, for each extracted pair, we calculate its distance to all pairs from PDTB using the normalized Levenshtein distance: 

\begin{equation}
\begin{split}
    d(s_1, s_2) = \frac{\text{Levenshtein}(s_1, s_2)}{\max(\text{len}(s_1), \text{len}(s_2))}
\end{split}
\label{eq:leven}
\end{equation}

\subsection{Implicit vs. Explicit Prediction Task Setup}

For each pair of connected sentences, whose relation type has been labeled in PDTB, the discourse relation between them may have been explicitly marked (via a discourse relation word) or not.
We can pose the task of a binary classification of whether the sentence pair appeared as explicitly or implicitly marked, given only the two sentences and no additional information. 
We evaluate DisSent and InferSent sentence embedding models and a word vector baseline on this trask.

We follow \citet{patterson2013predicting}'s preprocessing. The dataset contains 25 sections in total. We use sections 0 and 1 as the development set, sections 23 and 24 for the test set, and we train on the remaining sections 2-22.

This task is different from the setting in \citet{patterson2013predicting}. We do not allow the classifier to access the underlying discourse relation type and we only provide the individual sentence embeddings as input features. In contrast, \citet{patterson2013predicting} used a variety of discrete features provided by the PDTB dataset for their classifier, including the hand-annotated relation types.

\subsection{Classification Performance \label{sec:balanced_confusion}}

\begin{figure}[htb]
    \centering
    \includegraphics[width=0.4\textwidth]{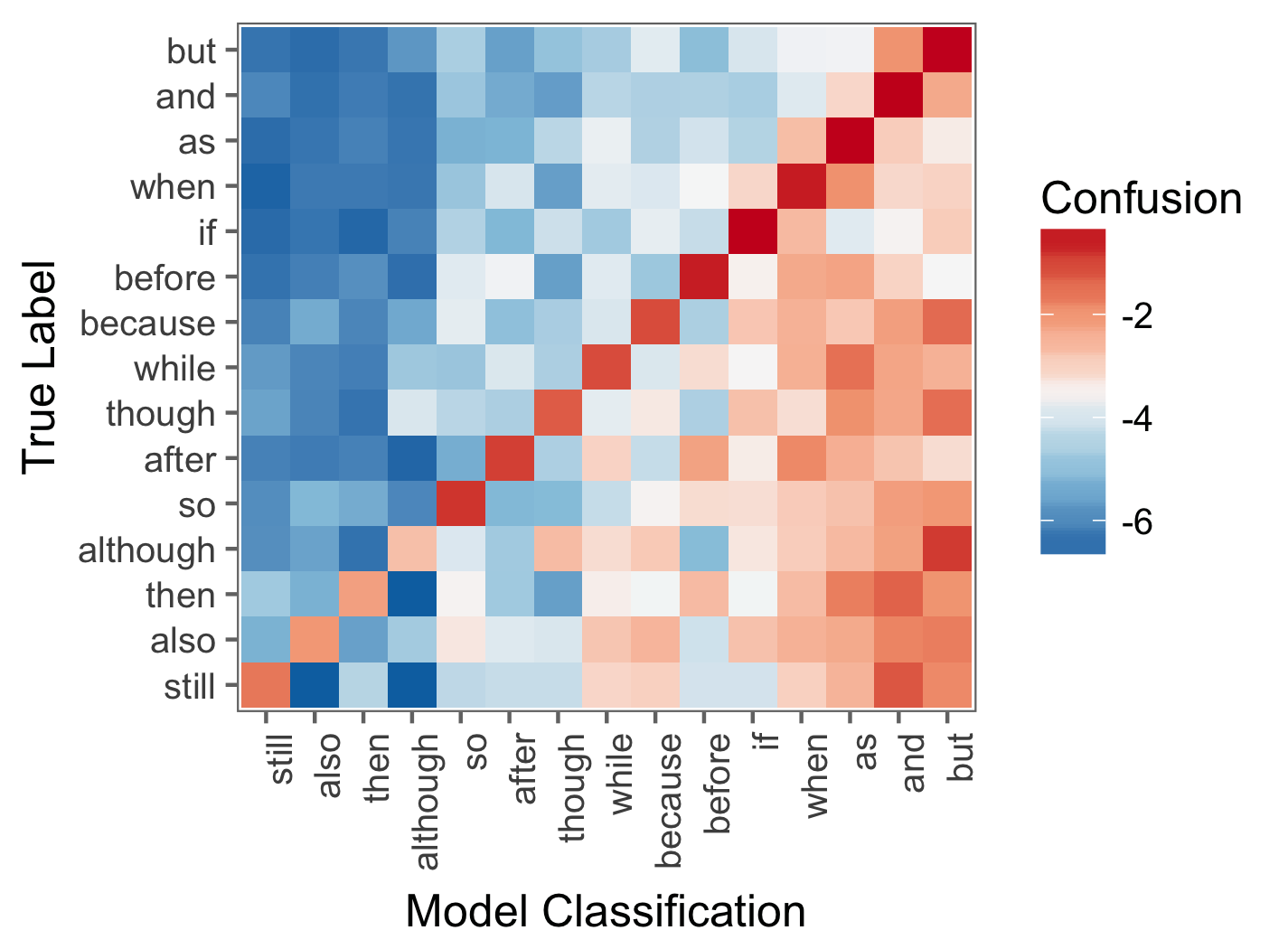}
    \caption{\textbf{Confusion Matrix} trained on the \ds{All} dataset extracted from BookCorpus. 
    Each cell represents the proportion of instances of the actual discourse marker misclassified as the classified discourse marker.
    This proportion is log-transformed to highlight small differences.
    Discourse markers are arranged in order of frequency from left (least frequent) to right (most frequent).}
    \label{fig:confusion}
\end{figure}

To investigate the qualitative relations among our largest set of discourse markers, the \ds{All} marker set, we build a confusion matrix of the test set classifications.
Figure \ref{fig:confusion} reflects classification performance for the model trained on the full dataset, that we later show generalization results for.
% Overall, this model achieved 67.5\% accuracy on the classification task.
% This model shows a clear effect of frequency
This model is clearly influenced by frequency, such that it tends to misclassify infrequent discourse markers as frequent ones.
However, deviations from the effect of frequency appear to be semantically meaningful.
% We provide a more detailed analysis of the confusability of semantically similar discourse markers in the Appendix.
% We provide a more detailed analysis of the semantic information implicit in the representations
% cluster 

Classifications errors are much more common for semantically similar discourse marker pairs than would be expected from frequency alone.
The most common confusion is when the synonymous marker \w{although} is mistakenly classified as \w{but}.
The temporal relation markers \w{before}, \w{after} and \w{then}, intuitively very similar discourse markers, are rarely confused for anything but each other.
The fact that they are indeed confusable may reflect the tendency of authors to mark temporal relation primarily when it is ambiguous.

\begin{figure}[htb]
    \centering
    \includegraphics[width=0.4\textwidth]{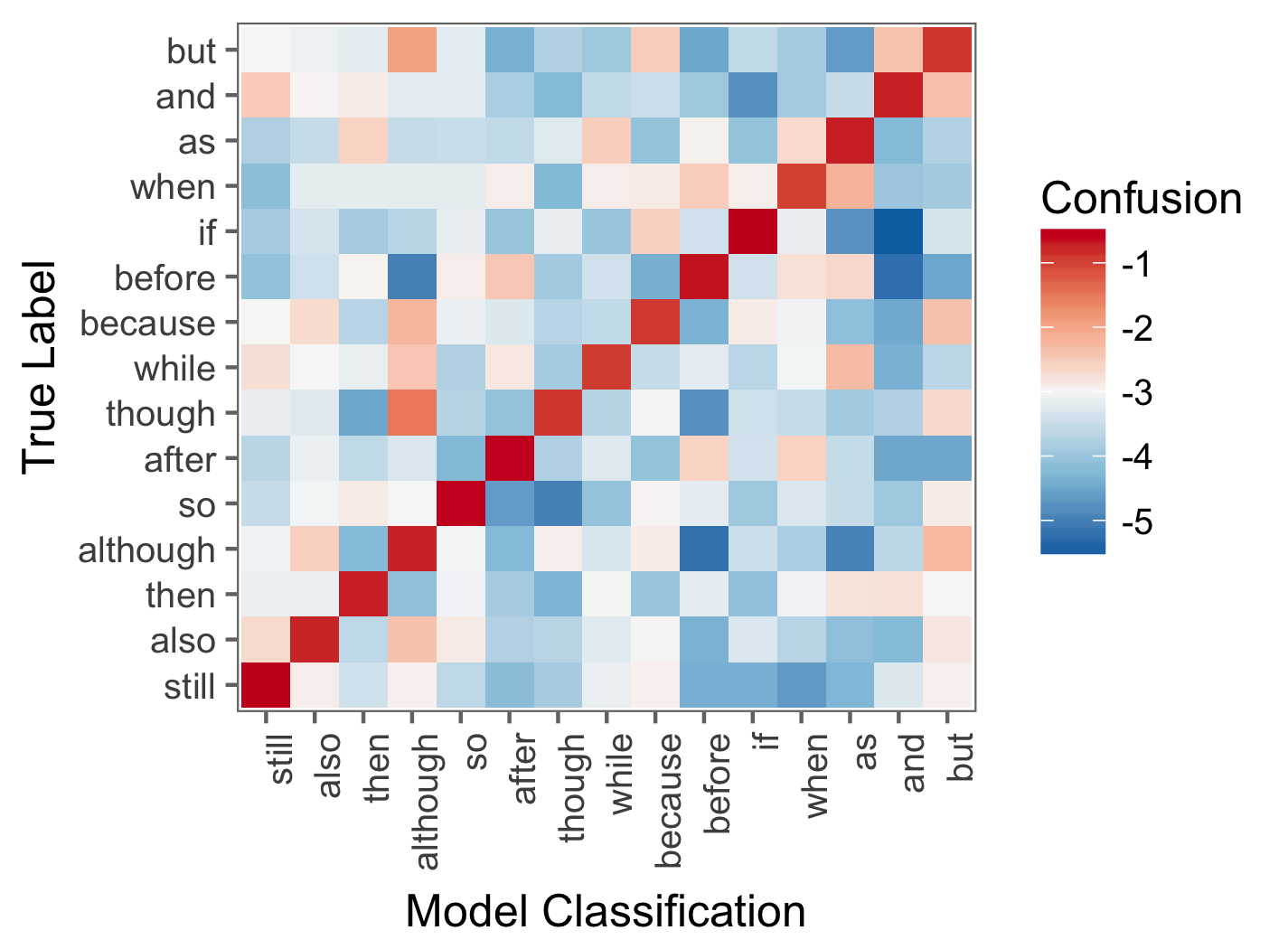}
    \caption{\textbf{Balanced Classifier Confusion Matrix} trained on a balanced subset of the \ds{All} dataset where discourse markers are capped at 13,421 occurrences each. 
    Each cell represents the proportion of instances of the actual discourse marker misclassified as the classified discourse marker.
    This proportion is log-transformed to highlight small differences.
    Discourse markers are arranged in order of frequency from left (least frequent) to right (most frequent).}
    \label{fig:balanced_confusion}
\end{figure}

Figure \ref{fig:balanced_confusion} reflects a model trained on a balanced subset of our training set.
When the model can no longer rely on base rates of discourse markers to make judgments, overall accuracy drops from 68\% to 47\%.
However inspecting the matrices shows very similar confusability, suggesting that training on unbalanced data does not greatly decrease sensitivity to non-frequency predictors.%

% \footnote{
To more quantitatively represent the connection between what the two models learn, we compute the correlation between the balanced confusions and the \emph{residuals} of the unbalanced confusions (when predicted linearly from log frequency).
% > cor(balanced_confusions$Confusion, balanced_confusions$resid)^2
% [1] 0.6430862
% > summary(lm(Confusion ~ resid, balanced_confusions))
%
% Call:
% lm(formula = Confusion ~ resid, data = balanced_confusions)
%
% Residuals:
%      Min       1Q   Median       3Q      Max 
% -1.58736 -0.38954 -0.00861  0.37621  1.80946 
%
% Coefficients:
%             Estimate Std. Error t value Pr(>|t|)    
% (Intercept) -3.29096    0.03773  -87.22   <2e-16 ***
% resid        0.70570    0.03521   20.05   <2e-16 ***
% ---
% Signif. codes:  0 ‘***’ 0.001 ‘**’ 0.01 ‘*’ 0.05 ‘.’ 0.1 ‘ ’ 1
%
% Residual standard error: 0.566 on 223 degrees of freedom
% Multiple R-squared:  0.6431,	Adjusted R-squared:  0.6415 
% F-statistic: 401.8 on 1 and 223 DF,  p-value: < 2.2e-16
These residuals account for 64\% of the variance in the balanced confusions ($R^2=0.6431, F(1,223)=401.8, p < .001$).
That is, we come close to predicting the balanced confusions from the unbalanced ones.% (Figure \ref{fig:residuals}).
%  Df  Sum Sq Mean Sq F value    Pr(>F)  
%   1 126.947 126.947  414.84 < 2.2e-16 ***
%  R2 = .12, F(1, 225) = 42.64, p < .001.
% }

\subsection{Baseline performance on training task\label{sec:intrinsic_baselines}}

As a reference point %to evaluate 
for training task performance %of our DisSent model, 
we present baseline performance.
Note that a model which simply chose the most common class would perform with 21.79\% accuracy on the \ds{ALL} task, 28.35\% on the \ds{Books 8} task, and 31.87\% on the \ds{Books 5} task.
Using either unigram, bigram and trigram bag of words or %either bag of words\footnote{For the bag of words model, we took unigram, bigram, and trigram frequencies for words that appeared at least 3 times in the corpus. This resulted in 3102426 ngram dimensions for \ds{Books 5}, 3397359 ngram dimensions for \ds{Books 8}, and 4275536 ngram dimensions for \ds{ALL}.} or smoothed average word embeddings
\citet{arora2016simple}'s baseline sentence representations as features to a logistic regression
% \footnote{For the Arora et al model, we the full dataset to estimate the word probabilities $p(w)$ and the first principle component $c_0$. We computed sentence vectors for the two sentences separately and concatenated the representation into a 600-dimensional vector for each sentence pair. We used $a=10^-3$, as in \cite{arora2016simple}'s simulations.} for each sentence as features in logistic regression 
results in much lower performance than our DisSent classifier.
Table \ref{table:intrinsic_bagowords} shows the precision and recall for the bag-of-words model.
Table \ref{table:intrinsic_avg} shows the precision and recall for the \citet{arora2016simple} embeddings.

\begin{table}[htb]
\centering
\footnotesize
\begin{tabular}{ c | c c c c c c}
\toprule
 & \multicolumn{2}{c}{All} & \multicolumn{2}{c}{Books 8} & \multicolumn{2}{c}{Books 5} \\
Marker & Prec & Rec & Prec & Rec & Prec & Rec \\
\midrule
and & 71.8 & 78.2 & 78.3 & 78.5 & 80.6 & 79.4 \\ 
but & 71.4 & 73.2 & 72.3 & 79.1 & 75.3 & 79.9 \\ 
because & 44.9 & 36.2 & 50.1 & 36.9 & 54.8 & 37.7 \\ 
if & 79.1 & 75.0 & 77.5 & 79.6 & 80.7 & 81.4 \\ 
when & 60.5 & 61.8 & 71.2 & 74.0 & 76.9 & 77.2 \\ 
so & 49.3 & 48.0 & 55.8 & 46.1 & --- & --- \\ 
though & 48.0 & 29.7 & 61.0 & 38.8 & --- & --- \\ 
before & 65.0 & 60.9 & 76.6 & 63.5 & --- & --- \\ 
as & 68.0 & 76.5 & --- & --- & --- & --- \\ 
while & 45.6 & 35.9 & --- & --- & --- & --- \\ 
after & 55.5 & 41.9 & --- & --- & --- & --- \\ 
although & 24.4 & 6.7 & --- & --- & --- & --- \\ 
still & 42.0 & 20.9 & --- & --- & --- & --- \\ 
also & 36.1 & 13.6 & --- & --- & --- & --- \\ 
then & 30.9 & 11.7 & --- & --- & --- & --- \\ 
\midrule 
Avg & 66.7 & 68.0 & 73.6 & 73.3 & 77.5 & 77.4 \\ 
\midrule
Accuracy & \multicolumn{2}{c}{67.5} & \multicolumn{2}{c}{73.5} & \multicolumn{2}{c}{77.3} \\
\bottomrule
\end{tabular}
\caption{\textbf{DisSent model performance:} Test recall / precision for each discourse marker on the classification task, weighted average precision and recall across discourse markers, and overall accuracy.}
 \label{table:dissent_intrinsic_eval}
\end{table}

% 52.833333333333336
% 44.68
% 67.85
% 62.0625
% 73.66
% 71.11999999999999
\begin{table}[htb]
\centering
\footnotesize
\begin{tabular}{ c | c c c c c c}
\toprule
 & \multicolumn{2}{c}{All} & \multicolumn{2}{c}{Books 8} & \multicolumn{2}{c}{Books 5} \\
Marker & Prec & Rec & Prec & Rec & Prec & Rec \\
\midrule
and      & 60.1 & 65.0 &    65.6 & 70.1 &    70.1 & 71.4 \\
but      & 49.9 & 65.3 &    55.2 & 69.4 &    59.7 & 69.9 \\
because  & 34.7 & 10.2 &    42.1 & 11.1 &    42.8 & 10.6 \\
if       & 54.6 & 56.9 &    58.8 & 56.4 &    64.4 & 60.0 \\
when     & 43.2 & 40.1 &    52.1 & 52.2 &    58.4 & 54.3 \\
so       & 35.5 & 11.3 &    38.5 & 11.0 &    --- & --- \\
though   & 40.6 & 20.8 &    56.2 & 25.2 &    --- & --- \\
before   & 47.8 & 29.1 &    56.6 & 35.4 &    --- & --- \\
as       & 51.9 & 63.1 &    --- & --- &    --- & --- \\
while    & 33.4 & 11.6 &    --- & --- &    --- & --- \\
after    & 41.0 & 17.6 &    --- & --- &    --- & --- \\
although & 11.9 &  0.4 &    --- & --- &    --- & --- \\
still    & 34.7 &  2.5 &    --- & --- &    --- & --- \\
also     & 16.7 &  0.4 &    --- & --- &    --- & --- \\
then     & 36.2 &  2.1 &    --- & --- &    --- & --- \\
\midrule
Average & 40.2 & 40.3 & 46.2 & 44.5 & 53.3 & 50.7 \\ 
\midrule
Accuracy &
  \multicolumn{2}{c}{51.8} & 
  \multicolumn{2}{c}{58.1} & 
  \multicolumn{2}{c}{63.3} \\
\bottomrule
\end{tabular}
\caption{\textbf{Ngram Bag-of-words baseline sentence embeddings performance on DisSent training task:} test recall / precision for each discourse marker on the classification task, and overall accuracy. Average metric reports the weighted average of all classes.}
 \label{table:intrinsic_bagowords}
\end{table}

%%% Ummmmm..... I need to check this implementation, because the Arora model should be doing better than this. :(
%%% Or wait... maybe not...? DisSent has a more complicated classifier... maybe I should give the model the multiplication, too...

% before: precision=48.80, recall=11.75
% though: precision=29.74, recall=1.33

\begin{table}[htb]
\centering
\footnotesize
\begin{tabular}{ c | c c c c c c}
\toprule
 & \multicolumn{2}{c}{All} & \multicolumn{2}{c}{Books 8} & \multicolumn{2}{c}{Books 5} \\
Marker & Prec & Rec & Prec & Rec & Prec & Rec \\
\midrule
and      & 46.9 & 59.4 &    52.9 & 63.6 &    58.0 & 64.3 \\
but      & 38.1 & 57.9 &    43.5 & 62.3 &    48.9 & 62.4 \\
because  & 24.1 &  0.5 &    20.2 & 0.3 &    27.7 & 0.47 \\
if       & 41.8 & 37.1 &    46.2 & 37.9 &    50.5 & 38.2 \\
when     & 36.8 & 25.8 &    45.6 & 40.0 &    58.3 & 41.3 \\
so       & 37.0 &  2.5 &    39.5 &  2.9 &    --- & --- \\
though   & 27.2 &  1.4 &    29.7 &  1.3 &    --- & --- \\
before   & 42.0 & 10.0 &    48.8 & 11.8 &    --- & --- \\
as       & 43.4 & 55.6 &     --- &  --- &    --- & --- \\
while    & 29.1 &  3.4 &     --- &  --- &    --- & --- \\
after    & 37.1 &  4.8 &     --- &  --- &    --- & --- \\
although &  0.0 &  0.0 &     --- &  --- &    --- & --- \\
still    &  0.0 &  0.0 &     --- &  --- &    --- & --- \\
also     &  0.0 &  0.0 &     --- &  --- &    --- & --- \\
then     &  0.0 &  0.0 &     --- &  --- &    --- & --- \\
\midrule
Avg & 50.1 & 51.1 & 57.5 & 56.4 & 63.0 & 62.2 \\
\midrule
Accuracy &
  \multicolumn{2}{c}{41.8} & 
  \multicolumn{2}{c}{47.3} & 
  \multicolumn{2}{c}{52.5} \\
\bottomrule
\end{tabular}
\caption{\textbf{Corrected GloVe Bag-of-words sentence embeddings performance on DisSent training task:} test recall / precision for each discourse marker on the classification task, and overall accuracy. Average metric reports the weighted average of all classes.}
 \label{table:intrinsic_avg}
\end{table}

\subsection{Embedding dimensions of models}

DisSent uses a BiLSTM encoder with 4096 hidden state dimensions. InferSent \protect\cite{conneau2017supervised} uses 4096 embedding dimensions. Disc BiGRU \protect\cite{jernite_discourse-based_2017} hidden state has 512 dimensions. FastSent and FastSent + AE \protect\cite{hill2016learning} have 500 dimensions. SkipThought \protect\cite{kiros2015skip} and SkipThought-LN \protect\cite{conneau2017supervised} models trained on 600-dimension word embeddings and produced 2400-dimension sentence embeddings. DictRep (bow) is from \protect\citet{conneau2017supervised}. LSMTL \protect\cite{subramanian2018learning} uses 2048-dimension bi-directional GRU as encoder, and trained on 512 dimension word embeddings.

\subsection{Limitations of evaluation}

The generalization tasks that we (following \citet{conneau2017supervised}) use to compare models focus on sentiment, entailment, and similarity. 
These are narrow operational definitions of semantic meaning.
A model that generates meaningful sentence embeddings should excel at these tasks.  
However, success at these tasks does not necessarily imply that a model has learned a deep semantic understanding of a sentence. 
% Recently a wider variety of linguistically complex tasks have been proposed in an attempt to ameliorate this issue \cite{wang2018glue}.

Sentiment classification, for example, in many cases only requires the model to understand local structures. 
Text similarity can be computed with various textual distances (e.g., Levenshtein or Jaro distance) on bag-of-words, without a compositional representation of the sentence. 
Thus, the ability of our, and other, models to achieve high  performance on these metrics may reflect a competent representation sentence meaning; but more rigorous tests are needed to understand whether these embeddings capture sentence meaning in general.

% \section{Appendices}
% \label{sec:appendix}
% Appendices are material that can be read, and include lemmas, formulas, proofs, and tables that are not critical to the reading and understanding of the paper. 
% Appendices should be {\bf uploaded as supplementary material} when submitting the paper for review. Upon acceptance, the appendices come after the references, as shown here. Use
% \verb|\appendix| before any appendix section to switch the section
% numbering over to letters.

\end{document}